\documentclass[11pt]{article}
\usepackage{emnlp2016}
\usepackage{times}
\usepackage{url}
\usepackage{latexsym}
\usepackage{xspace}
\usepackage{booktabs}
\usepackage{color}
\usepackage{graphicx}
\usepackage{tabulary}
\usepackage{enumitem}

\usepackage{amsfonts}

  {\begin{itemize}[topsep=0pt, partopsep=0pt] %
    \setlength{\itemsep}{0pt}%
    \setlength{\parskip}{0pt}%
    }%
  {\end{itemize}}
  
  \usepackage{color}
  {\begin{enumerate}â \footnotesize%
    \setlength{\itemsep}{0pt}%
    \setlength{\parskip}{0pt}%
    }%
  {\end{enumerate}}

\makeatletter
\newcommand{\@BIBLABEL}{\@emptybiblabel}
\newcommand{\@emptybiblabel}[1]{}
\makeatother
\usepackage[hidelinks]{hyperref}

\usepackage{microtype}
\usepackage{multirow}
\usepackage{verbatim}
\usepackage{amsmath,amsthm,amssymb}
\usepackage{array}
\usepackage[scaled=0.86]{helvet}
\usepackage{ifthen}
\usepackage{courier}
\usepackage[linesnumbered,vlined,ruled]{algorithm2e}

\belowdisplayskip \abovedisplayskip

\newcommand{\sts}{{{\textsc{Seq2Seq}}}\xspace}

\title{Deep Reinforcement Learning for Dialogue Generation}
\author{Jiwei Li$^1$, Will Monroe$^1$, Alan Ritter$^2$, Michel Galley$^3$, Jianfeng Gao$^3$ and Dan Jurafsky$^1$ \\
$^1$Stanford University, Stanford, CA, USA\\
$^2$Ohio State University, OH, USA \\
$^3$Microsoft Research, Redmond, WA, USA \\
{\tt\{jiweil,wmonroe4,jurafsky\}@stanford.edu, ritter.1492@osu.edu} \\
{\tt\{mgalley,jfgao\}@microsoft.com}
}
\date{}
\emnlpfinalcopy
\begin{document}
\maketitle
\begin{abstract}
Recent neural models of dialogue generation offer great promise for
generating responses for conversational agents, but tend to be
shortsighted, predicting utterances one at a time while ignoring their influence on future outcomes.
Modeling the future direction of a dialogue is
crucial to generating coherent, interesting dialogues, a need which led
traditional NLP models of dialogue to draw on reinforcement learning.
In this paper, we show how to integrate these goals,
applying deep reinforcement learning to model future reward in chatbot dialogue.
The model simulates dialogues between two virtual agents, using policy gradient methods to reward sequences that display
three useful conversational properties: informativity, coherence, and ease of answering (related to forward-looking function).
We evaluate our model on diversity, length
as well as with human judges,  showing that the proposed algorithm generates more interactive responses and manages to foster a more sustained conversation in dialogue simulation. 
This work marks a first step towards learning a neural conversational model based on the long-term success of dialogues.
\end{abstract}

\section{Introduction}
\label{sec:intro}
Neural response generation \cite{sordoni2015neural,shang2015neural,vinyals2015neural,li2015diversity,wen2015semantically,yao2015attention,luan2016lstm,xu2016incorporating,wen2016network,li2016persona,Su} is of growing interest. The LSTM sequence-to-sequence (\sts) model \cite{sutskever2014sequence}  is one type of neural generation model that
maximizes the probability of generating a response given the previous dialogue turn.
This approach enables the incorporation of rich context when mapping between consecutive dialogue turns \cite{sordoni2015neural} in a way not possible, for example, 
with MT-based dialogue models \cite{ritter2011data}.

Despite the success of \sts models in dialogue generation, two problems emerge:
First, \sts models are trained by predicting the next dialogue turn in a given conversational context using the maximum-likelihood estimation (MLE) objective function. 
However, it is not clear how well MLE approximates the real-world goal of chatbot development:
teaching a machine to converse with humans, while providing interesting, diverse, and informative feedback that keeps users engaged.
One concrete example is that    \sts models tend to generate highly generic responses such as ``{\it I don't know}'' regardless of the input \cite{sordoni2015neural,serban2015hierarchical,li2015diversity}. This can be ascribed to the high frequency of generic responses found in the training set and their 
compatibility with a diverse range of conversational contexts.
Yet {\it ``I don't know''} is apparently not a good action to take, since it closes the conversation down. 

Another common problem, illustrated in the two sample conversations on the  left of Table 1, is that the system becomes stuck in an infinite loop of repetitive responses. This is due to MLE-based \sts models' inability to account for repetition.
In example 2 (bottom left), the dialogue falls into an infinite loop after three turns, with both agents generating dull, generic utterances like {\it i don't know what you are talking about} and {\it you don't know what you are saying}. 
Looking at the entire conversation, utterance (4) turns out to be a bad action to take
because it offers no way of continuing the conversation.\footnote{A similar rule is often suggested in improvisational comedy: \url{https://en.wikipedia.org/wiki/Yes,_and...}}
\begin{table*}
\centering
\small
\begin{tabular}{l|l}\hline
{\bf Baseline mutual information model (Li et al. 2015)}  & {\bf Proposed reinforcement learning model} \\\hline
\textsc{A}: {Where are you going?} (1)             & A: Where are you going? (1)   \\
\textsc{B}: {I'm going to the restroom}. (2)    & B: I'm going to the police station. (2) \\
\textsc{A}: {See you later}. (3)			    & A: I'll come with you. (3) \\
\textsc{B}: {See you later}. (4)			    & B: No, no, no, no, you're not going anywhere. (4) \\
\textsc{A}: {See you later}. (5)		    & A: Why? (5) \\
\textsc{B}: {See you later}. (6)			    & B: I need you to stay here. (6)\\
...	&								    A: I don't know what you are talking about. (7)\\
...& ... \\\hline
\textsc{A}: {how old are you?} (1) & A: How old are you? (1)\\
\textsc{B}: {I'm 16.} (2)& B: I'm 16. Why are you asking? (2)\\
\textsc{A}: {16?} (3) & A I thought you were 12. (3) \\
\textsc{B}: {I don't know what you are talking about.} (4) & B: What made you think so? (4)\\
\textsc{A}: {You don't know what you are saying.} (5) & A: I don't know what you are talking about. (5)\\
\textsc{B}: {I don't know what you are talking about .} (6) & B: You don't know what you are saying. (6)\\
\textsc{A}: {You don't know what you are saying.} (7) & ... \\
... & ... \\\hline
\end{tabular}					
\caption{{\bf Left Column}: Dialogue simulation between two agents using  a 4-layer LSTM encoder-decoder trained on the OpenSubtitles dataset. The first turn (index 1) is input by the authors. Then the two agents take turns conversing, taking as input the other agent's prior generated turn. The output is generated using the mutual information model (Li et al., 2015) in which an N-best list is first obtained using beam search based on p(t$|$s) and reranked by linearly combining the backward probability p(s$|$t), where t  and s respectively denote targets and  sources. {\bf Right Column}: Dialogue simulated using the proposed reinforcement learning model. The new model has more forward-looking utterances (questions like ``Why are you asking?" and offers like ``I'll come with you") and lasts longer before it falls into conversational black holes.}
\label{con-simulation}	
\end{table*}

These challenges suggest we need a conversation framework that has the ability to (1)  integrate developer-defined rewards that better mimic the true goal of chatbot development
  and (2)  model the long-term influence of a generated response in an ongoing dialogue.

To achieve these goals,  we draw on the insights of reinforcement learning, which
have been widely applied in MDP and POMDP dialogue systems (see Related Work section for details).
We introduce a neural reinforcement learning (RL) generation method, 
which can optimize long-term rewards designed by system developers.
Our model uses the encoder-decoder architecture as its backbone, and 
 simulates conversation between two virtual agents to explore the space of possible actions while learning to maximize expected reward.
We define simple heuristic approximations to rewards that characterize
 good conversations: good conversations are forward-looking \cite{All92} or interactive (a turn suggests
 a following turn), informative, and coherent.
  The parameters of an encoder-decoder RNN define a policy over an infinite action space consisting of all possible utterances.
  The agent learns a policy by optimizing the long-term developer-defined reward from ongoing dialogue simulations using policy gradient methods \cite{williams1992simple},
  rather than the MLE objective defined in standard \sts models. 

Our model thus integrates the power of \sts systems to learn compositional semantic
meanings of utterances with the strengths of reinforcement learning in
optimizing for long-term goals across a conversation.
 Experimental results (sampled results at the right panel of Table \ref{con-simulation}) demonstrate that our approach fosters a more sustained dialogue and
  manages to produce more interactive  responses than standard \sts models trained using the MLE objective.

\section{Related Work}
Efforts to build statistical dialog systems  fall into two major categories.

The first 
 treats dialogue generation as a source-to-target transduction problem 
and learns  mapping rules 
between input messages and responses from a massive amount of training data. \newcite{ritter2011data}
frames the response generation problem as a
statistical machine translation (SMT) problem.  
\newcite{sordoni2015neural}
improved Ritter et al.'s system 
by rescoring
the outputs of a phrasal SMT-based conversation
system with a neural model that incorporates
prior context.
Recent progress in \sts models inspire several efforts \cite{vinyals2015neural} to  build end-to-end conversational systems which first apply an encoder to map a message to a distributed vector representing its semantics and generate a response from the message vector. 
\newcite{serban2015hierarchical}  propose a hierarchical neural model that captures dependencies over an extended
conversation history. \newcite{li2015diversity} 
propose mutual information between message and
response as an alternative objective function in order to reduce the proportion of generic
responses produced by \sts systems.

The other line of statistical research focuses on building task-oriented dialogue systems to solve domain-specific tasks.  
Efforts include statistical models such as Markov Decision Processes (MDPs) \cite{levin1997learning,levin2000stochastic,walker2003trainable,pieraccini2009we},
POMDP \cite{young2010hidden,young2013pomdp,gavsic2013pomdp,gavsic2014incremental} models, and models that statistically learn generation
rules \cite{oh2000stochastic,ratnaparkhi2002trainable,banchs2012iris,nio2014developing}.
This dialogue literature thus widely applies reinforcement learning \cite{walker2000application,schatzmann2006survey,gasic2013line,singh1999reinforcement,singh2000empirical,singh2002optimizing}
to train dialogue policies.
But task-oriented RL dialogue systems often rely on carefully limited dialogue parameters,
or hand-built templates with state, action and reward signals designed by humans for each new domain,
making the paradigm difficult to extend to open-domain scenarios.

Also relevant is prior work on reinforcement learning for language understanding - including learning from delayed
reward signals by playing text-based games \cite{narasimhan2015language,he2015deep},
executing instructions for Windows help  \cite{branavan2011learning}, or understanding dialogues 
that give navigation directions \cite{vogel10}.

Our goal is to integrate the \sts and reinforcement learning  paradigms, drawing on the advantages of both.
We are thus particularly inspired by recent work that attempts to merge these paradigms, including
 \newcite{wen2016network}---
training an end-to-end task-oriented dialogue system that links input representations to slot-value pairs in a database---
or  \newcite{Su}, who combine reinforcement learning with neural generation on tasks with real users, showing that reinforcement learning improves dialogue performance.

\section{Reinforcement Learning for Open-Domain Dialogue}
In this section, we describe in detail the components of the proposed RL model. 

The learning system consists of two agents. 
We use $p$ to denote sentences generated from the first agent and  $q$ to denote sentences from the second. The two agents take turns talking with each other. A dialogue can be represented as an alternating sequence of sentences generated by the two agents: $p_1, q_1, p_2, q_2, ..., p_{i}, q_{i}$.  We view the generated sentences as actions that are taken according to a policy defined by an
encoder-decoder recurrent neural network language model.

The parameters of the network are optimized to maximize the expected future reward using policy search, as described in Section \ref{sec:learning}.  
Policy gradient methods are more appropriate for our scenario than Q-learning \cite{mnih2013playing}, because we can initialize the encoder-decoder RNN using
MLE parameters that already produce plausible responses, before changing the objective and tuning towards a policy that maximizes long-term reward.
Q-learning, on the other hand, directly estimates the future expected reward of each action, which can differ from the MLE objective by orders of magnitude,
thus making MLE parameters inappropriate for initialization.
The components (states, actions, reward, etc.) of our sequential decision problem are summarized in the following sub-sections.

\subsection{Action} An action $a$ is the dialogue utterance to generate. The action space is infinite since arbitrary-length sequences can be generated.

\subsection{State} A state is denoted by the previous two dialogue turns $[p_i,q_i]$. 
The dialogue history is further transformed to a vector representation by feeding the concatenation of $p_i$ and $q_i$ into an LSTM encoder model as described in \newcite{li2015diversity}.

\subsection{Policy} A policy takes the form of an LSTM encoder-decoder (i.e., $p_{RL}(p_{i+1}|p_{i}, q_i)$ ) and is defined by its parameters.
Note that we use a stochastic representation of the policy (a probability distribution over actions given states).
A deterministic policy would result in a discontinuous objective that is difficult to optimize using gradient-based methods.

\subsection{Reward} $r$ denotes the reward obtained for each action. In this subsection, we discuss major factors that contribute to the success of a dialogue and describe how 
approximations to these factors can be operationalized in computable reward functions. 

\paragraph{Ease of answering}
A turn generated by a machine should be easy to respond to.
This aspect of a turn is related to its {\em forward-looking function}: the constraints
a turn places on the next turn \cite{schegloff73,All92}.
We propose to measure the ease of answering a generated turn by
using the negative log likelihood of responding to that utterance with a dull response.
We manually constructed a list of dull responses $\mathbb{S}$ consisting 8 turns such as ``I don't know what you are talking about", ``I have no idea", etc., that we and others have found
occur very frequently in \sts models of conversations.
The reward function is given as follows:
\begin{equation}
r_1=-\frac{1}{N_{\mathbb{S}}}\sum_{s\in \mathbb{S}}\frac{1}{N_s}\log p_{\text{seq2seq}} (s|a)
\label{eq1}
\end{equation}
where $N_{\mathbb{S}}$ denotes the cardinality of $N_{\mathbb{S}}$ and $N_s$ denotes the number of tokens in the dull response $s$. 
Although of course there are more ways to generate dull responses than the list can cover,
many of these responses are likely to fall into similar regions in the vector space computed by the model.
A system less likely to generate utterances in the list is thus also less likely to generate other dull responses. 

$p_{\text{seq2seq}}$ represents the  likelihood output by \sts models. 
It is worth noting that $p_{\text{seq2seq}}$ is different from the stochastic policy function $p_{RL}(p_{i+1}|p_{i}, q_i)$, since the former is learned based on the MLE objective of the \sts model while the latter is the policy optimized for long-term future reward in the RL setting.
$r_1$ is further scaled by the length of target $\mathbb{S}$.
\paragraph{Information Flow}
We want each agent to contribute new information at each turn to keep the dialogue moving and avoid repetitive sequences. 
We therefore propose penalizing semantic similarity between consecutive turns from the same agent. 
Let $h_{p_i}$ and $h_{p_{i+1}}$ denote representations obtained from the encoder for two consecutive turns $p_i$ and $p_{i+1}$. The reward is given by the negative log of the cosine similarity
between them:
\begin{equation}
r_2=-\log \text{cos}(h_{p_i},h_{p_{i+1}}) = -\log \text{cos} \frac{h_{p_i} \cdot h_{p_{i+1}}}{\lVert h_{p_i} \rVert \lVert h_{p_{i+1}} \rVert}
\label{sim}
\end{equation}
\paragraph{Semantic Coherence}
We also need to measure the adequacy of responses to avoid situations in which the generated replies are highly rewarded but are ungrammatical or not coherent.
 We therefore  consider the mutual information between the action $a$ and previous turns in the history to ensure the generated responses are coherent and appropriate: 
\begin{equation}
r_3=\frac{1}{N_a}\log p_{\text{seq2seq}}(a|q_i,p_i)+\frac{1}{N_{q_i}}\log p_{\text{seq2seq}}^{\text{backward}}(q_i|a)
\label{eq4}
\end{equation}
$p_{\text{seq2seq}}(a|p_i,q_i)$  denotes the probability of generating response $a$ given the previous dialogue utterances $[p_i,q_i]$. $p_{\text{seq2seq}}^{\text{backward}}(q_i|a)$ denotes the backward probability of generating the previous dialogue utterance $q_i$ based on response $a$. $p_{\text{seq2seq}}^{\text{backward}}$ is trained in a similar way as standard \sts models with sources and targets swapped. 
Again, to control the influence of target length, both $\log p_{\text{seq2seq}}(a|q_i,p_i)$ and  $\log p_{\text{seq2seq}}^{\text{backward}}(q_i|a)$ are scaled by the length of targets.

The final reward for  action $a$ is a weighted sum of the rewards discussed above:
\begin{equation}
r(a,[p_i,q_i])=\lambda_1 r_1+\lambda_2 r_2+\lambda_3 r_3
\label{reward}
\end{equation}
where $\lambda_1+\lambda_2+\lambda_3=1$.
We set $\lambda_1=0.25$, $\lambda_2=0.25$ and $\lambda_3=0.5$. 
A reward is observed after the agent  reaches the end of each sentence.

\section{Simulation}
The central idea behind our approach is to simulate the process of two virtual agents taking turns talking with each other, through which we can explore the state-action space and learn a policy
$p_{RL}(p_{i+1}|p_{i}, q_i)$ that leads to the optimal expected reward.
We adopt an AlphaGo-style strategy \cite{silver2016mastering} by initializing the RL system using a general response generation policy which is learned from a fully supervised setting. 
\subsection{Supervised Learning}
 For the first stage of training, we build on prior work of predicting a generated target sequence given dialogue history using the supervised \sts model \cite{vinyals2015neural}.
 Results from supervised models will be later used for initialization. 
 
We trained a \sts model with attention \cite{bahdanau2014neural} on the OpenSubtitles dataset, which consists of roughly 80 million source-target pairs. 
We treated each turn in the dataset as a target and the concatenation of two previous sentences as source inputs. 
\subsection{Mutual Information}
Samples from \sts models are often times dull and generic, e.g., ``{\it i don't know}" \cite{li2015diversity}
We thus do not want to initialize the policy model using the pre-trained \sts models because this will lead to a lack of diversity in the RL models' experiences. 
 \newcite{li2015diversity} showed that modeling mutual information between sources and targets will significantly decrease the chance of generating dull responses and improve  general response quality.
We now show how we can obtain an encoder-decoder model which generates maximum mutual information responses.

As illustrated in \newcite{li2015diversity}, direct decoding from Eq~\ref{eq4} is infeasible since the second term 
 requires
 the target sentence to be  completely generated.
 Inspired by recent work on sequence level learning \cite{ranzato2015sequence}, we treat the problem of 
generating maximum mutual information response 
as a reinforcement learning problem in which a reward of mutual information value is observed when the model arrives at the end of  a sequence. 

Similar to \newcite{ranzato2015sequence}, we use policy gradient methods \cite{sutton1999policy,williams1992simple} for optimization. We initialize the policy model $p_{RL}$ using a pre-trained $p_{\sts}(a|p_i,q_i)$ model. 
Given an input source $[p_i,q_i]$, 
we generate a candidate list  $A=\{\hat{a}|\hat{a}\sim p_{RL}\}$. 
For each generated candidate $\hat{a}$, we will obtain the mutual information score $m(\hat{a}, [p_i,q_i])$ from the pre-trained $p_{\sts}(a|p_i,q_i)$ and $p^{\text{backward}}_{\sts}(q_i|a)$.
This mutual information score will be used as a reward and back-propagated to the encoder-decoder model, tailoring it to generate sequences with higher rewards. 
We refer the readers to \newcite{zaremba2015reinforcement}  and
\newcite{williams1992simple} for details.  The expected reward for a sequence is given by:
\begin{equation}
\begin{aligned}
&J(\theta)=\mathbb{E} [m(\hat{a},[p_i,q_i]) ]\\
\end{aligned}
\end{equation}
The gradient is
estimated using the likelihood ratio trick:
\begin{equation}
\begin{aligned}
\nabla J(\theta)=m(\hat{a},[p_i,q_i])\nabla\log p_{RL}(\hat{a}|[p_i,q_i])
\end{aligned}
\end{equation}
We update the parameters in the encoder-decoder model using stochastic gradient descent. 
A
curriculum learning strategy is adopted \cite{bengio2009curriculum} 
as in \newcite{ranzato2015sequence}  
such that, for
every sequence of length $T$ we use the MLE loss for the first $L$ tokens and the reinforcement algorithm for the remaining $T-L$ tokens. We gradually anneal 
the value of $L$ to zero.
A baseline strategy is employed to decrease the learning variance:
an additional neural model takes as inputs the generated target and the initial source and outputs a baseline value, similar to the strategy 
adopted by \newcite{zaremba2015reinforcement}. 
 The final gradient is thus:
\begin{equation}
\begin{aligned}
\nabla J(\theta)=\nabla\log p_{RL}(\hat{a}|[p_i,q_i]) [m(\hat{a},[p_i,q_i])-b]
\end{aligned}
\end{equation}

\subsection{Dialogue Simulation between Two Agents}
\label{sec:learning}
We simulate conversations between the two virtual agents and 
have them take turns talking with each other. The simulation proceeds as follows: at the initial step, a message from the training set is fed to the first agent. The agent encodes the input message to a vector representation and starts decoding to generate  a response output. 
Combining the immediate output from the first agent with the dialogue history, the second agent updates the state by encoding the dialogue history into a representation and uses the decoder RNN to generate responses, which are subsequently fed back to the first agent, and the process is repeated. 

\begin{figure*}[!ht]
    \centering
    \includegraphics[width=6in]{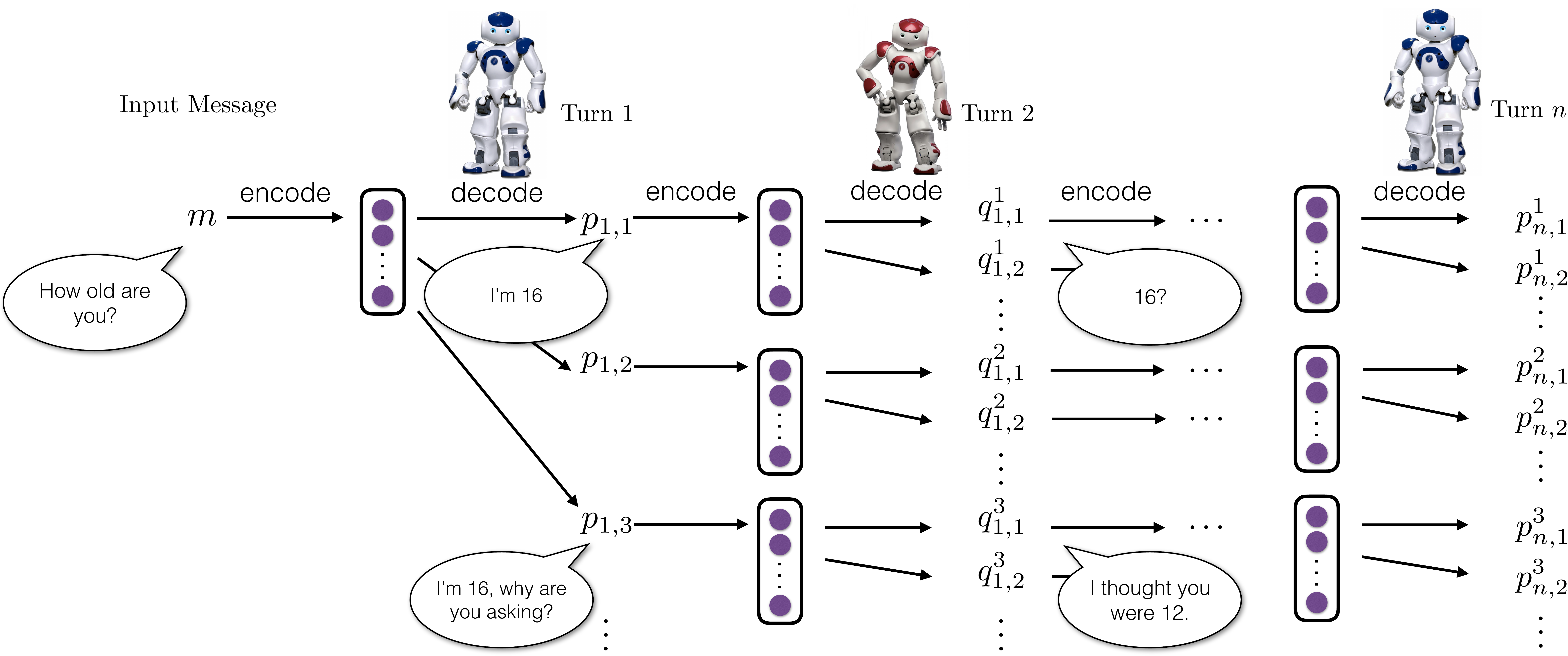}
\caption{Dialogue simulation between the two agents. }
\end{figure*}

\paragraph{Optimization}
We initialize the policy model $p_{RL}$ with parameters from the mutual information model described in the previous subsection. 
We then use policy gradient methods to find parameters that lead to a larger expected reward.
The objective to maximize is the expected future reward:
\begin{equation}
J_{RL}(\theta)=\mathbb{E}_{p_{RL}(a_{1:T})} [ \sum_{i=1}^{i=T} R(a_i,[p_i,q_i] ) ]
\end{equation}
where $R(a_i,[p_i,q_i])$ denotes the reward resulting from action $a_i$. 
We
use the likelihood ratio trick 
\cite{williams1992simple,glynn1990likelihood,aleksand1968stochastic} for gradient updates:
\begin{equation}
\nabla J_{RL}(\theta) \approx { \sum_{i}\nabla\log p(a_i|p_i,q_i)} \sum_{i=1}^{i=T} R(a_i,[p_i,q_i] )
\end{equation}

We  refer readers to
\newcite{williams1992simple} and \newcite{glynn1990likelihood} for more details. 
\subsection{Curriculum Learning}
A curriculum Learning strategy is again employed  in which we begin by simulating the dialogue for 2 turns, and gradually increase the number of simulated turns. 
We generate 5 turns at most, as  the number of candidates to examine grows exponentially in the size of candidate list.
Five candidate responses are generated at each step of the simulation.

\section{Experimental Results}
In this section, we  describe experimental results along with qualitative analysis.  We evaluate dialogue generation systems using both human judgments and
two automatic metrics: conversation length (number of turns in the entire session) and diversity.
\subsection{Dataset}
The dialogue simulation requires  high-quality initial inputs fed to the agent. For example, an initial input of ``why ?" is undesirable since it is unclear how the dialogue could proceed. 
We take a subset of 10 million messages from the OpenSubtitles dataset and extract 0.8 million sequences with the lowest likelihood of generating the response
``{\it i don't know what you are taking about}" to ensure initial inputs are easy to respond to.

\subsection{Automatic Evaluation}
Evaluating dialogue systems is difficult. Metrics such as BLEU
\cite{papineni2002bleu} and perplexity have been widely used for
dialogue quality evaluation
\cite{li2015diversity,vinyals2015neural,sordoni2015neural}, but it
is widely debated how well these automatic metrics are correlated
with true response quality \cite{liu2016not,galley2015deltableu}.
Since the goal of the proposed system is not to predict the highest probability response, but rather the long-term success of the dialogue, 
we do not employ BLEU or perplexity for evaluation\footnote{We found the RL model performs worse on BLEU score. On a random sample of 2,500 conversational pairs, single reference BLEU scores for RL models, mutual information models and vanilla \sts models  are respectively 1.28, 1.44 and 1.17.  BLEU is highly correlated with perplexity in generation tasks. Since the RL model is trained based on future reward rather than MLE, it is not surprising that the RL based models achieve lower BLEU score.}.

\paragraph{Length of the dialogue}
The first metric we propose is the length of the simulated dialogue. We say a dialogue ends when one of the agents starts generating dull responses such as ``{\it i don't know}" \footnote{We use a simple rule matching method, with a list of 8 phrases that count as dull responses.  Although this can lead to both false-positives and -negatives, it works pretty well in practice. }
or two consecutive utterances from the same user are highly overlapping\footnote{Two utterances are considered to be repetitive if they share more than 80 percent of their words.}.


The test set consists of 1,000 input messages. 
To reduce the risk of circular dialogues, we limit the number of simulated turns to be less than 8.  
Results are shown in Table \ref{numturns}. As can be seen, using mutual information leads to more sustained conversations between the two agents.
The proposed  RL model is first trained based on the mutual information objective and thus benefits from it in addition to the RL model. 
We observe that the RL model with dialogue simulation achieves the best evaluation score.

\begin{table}
\centering
\begin{tabular}{c|c}
Model& $\#$ of simulated turns \\\hline
\sts& 2.68 \\
mutual information& 3.40\\
RL&4.48 \\\hline
\end{tabular}
\caption{The average number of simulated turns from standard \sts models, mutual information model and the proposed RL model.}
\label{numturns}
\end{table}

\begin{table*}[!ht]
\centering
\small
\resizebox{\textwidth}{!}{
\begin{tabular}{lll}\hline
Input Message& Mutual Information Model& RL \\\hline
How old are you? & I'm 16. & I'm 16, why are you asking? \\
What's your full name? & I have no idea. & What's yours? \\
I don't want to go home tonight. & Really? & Why? \\
Do you have any feelings for me? & I don't know what you are talking about. &I don't want to hurt your feelings. \\
How much time do you have here? & Not long enough. Sorry, sir. & Ten seconds. \\
Shall we get started?& Of course! & Yes. We've got a lot of work to do here.\\
Do you play football? & No, i don't & Yes. I love football!\\ 
We'd have to talk to him.&I mean, he's a good guy&About what ? \\
How come you never say it? & Because I don't want to hurt you.& I don't think it's a good idea to say it. \\\hline
\end{tabular}
}
\caption{Sampled responses generated from the mutual information models and the proposed RL model.}
\label{sen}
\end{table*}

\paragraph{Diversity} 
We report degree of diversity by calculating
the number of distinct unigrams and bigrams
in generated responses. The value is scaled by the total
number of generated tokens to avoid favoring long
sentences as described in \newcite{li2015diversity}.
The resulting metric is thus a type-token ratio for unigrams and bigrams.

For both the standard \sts model and the proposed RL model, we use beam search with a beam size 10 to generate a response to a given input message. For the mutual information model, we first generate $n$-best lists using $p_{\sts}(t|s)$ and then linearly re-rank them using $p_{\sts}(s|t)$.
Results are presented in Table \ref{diversity}. We find that the proposed RL model generates more diverse outputs when compared against both the vanilla \sts model and the mutual information model. 

\begin{table}[ht]
\centering
\begin{tabular}{c|l|l}
Model& Unigram&Bigram \\\hline
\sts& 0.0062&0.015 \\
mutual information& 0.011&0.031\\
RL&0.017&0.041 \\\hline
\end{tabular}
\caption{Diversity scores (type-token ratios) for the standard \sts model, mutual information model and the proposed RL model.}
\label{diversity}
\end{table}

\paragraph{Human Evaluation}
We explore three settings for human evaluation:
the first setting is similar to what was described in \newcite{li2015diversity},
where we employ crowdsourced judges to evaluate a random sample of 500 items. 
We present both an input message and the generated outputs to 3 judges and ask them to 
 decide which of the two outputs is better (denoted as {\it single-turn general quality}). 
Ties are permitted. 
Identical strings are assigned the same score. 
We measure the improvement achieved by the RL model over the mutual information model by the mean difference in scores between the models. 

For the second setting, judges are again presented with input messages and system outputs, but are asked to decide which of the two outputs is easier to respond to (denoted as {\it single-turn ease to answer}).
Again we evaluate a random sample of 500 items, each being assigned to 3 judges.

For the third setting, judges are presented with simulated conversations between the two agents (denoted as {\it multi-turn general quality}). Each conversation consists of 5 turns. 
We evaluate 200 simulated conversations, each being assigned to 3 judges, who are asked to decide which of the simulated conversations is of higher quality.

\begin{table}[!ht]
\centering
\small
\begin{tabular}{cccc} \\\hline
Setting&RL-win&RL-lose&Tie  \\\hline
single-turn general quality&0.40&0.36&0.24 \\
single-turn ease to answer& 0.52&0.23&0.25\\
multi-turn general quality&0.72&0.12&0.16\\\hline
\end{tabular}
\caption{RL gains over the mutual information system based on pairwise human judgments.}
\label{human-eval}
\end{table}

Results for human evaluation are shown in Table \ref{human-eval}. 
The proposed RL system does not introduce a significant boost in single-turn response quality (winning 40 percent of time and losing 36 percent of time).
This is in line with our expectations, 
as the RL model is not optimized to predict the next utterance, but rather to increase long-term reward.
The RL system produces responses that are significantly easier to answer than 
does the mutual information system, as demonstrated by the {\it single-turn ease to answer} setting (winning 52 percent of time and losing 23 percent of time),
and also significantly higher quality multi-turn dialogues, as demonstrated by the {\it multi-turn general quality} setting (winning 72 percent of time).

\paragraph{Qualitative Analysis and Discussion}
We show a random sample of generated responses in Table \ref{sen} and simulated conversations in Table \ref{con-simulation} at the beginning of the paper.
From Table \ref{sen}, we can see that the RL based agent indeed generates more interactive responses than the other baselines. We also find that the RL model has a tendency to 
end a sentence with another question and hand the conversation over to the user. 
From Table \ref{con-simulation}, we observe that the RL model manages to produce more interactive and sustained conversations than the mutual information model. 

During error analysis, we found that although we penalize repetitive utterances in consecutive turns, the dialogue sometimes enters a cycle with length greater than one, as shown in Table \ref{he}. This can be ascribed to the 
limited amount of conversational history we consider. 
Another issue observed is that the model sometimes starts a less relevant topic during the conversation. There is a tradeoff between relevance and less repetitiveness, as manifested in the reward function we define in Eq~4. 

\begin{table}[!ht]
\centering
\small
\begin{tabular}{l}\hline
A: What's your name ?\\
B: Daniel.  \\
A: How old are you ?\\
B. Twelve. What's your name ? \\
A. Daniel.  \\
B: How old are you ? \\
A: Twelve. What's your name ? \\
B: Daniel.\\
A: How old are you ? \\
B ... \\\hline
\end{tabular}
\caption{An simulated dialogue with a cycle longer than one.}
\label{he}
\end{table}

The fundamental problem, of course, is that the manually defined reward function 
can't possibly cover the crucial aspects that define an ideal conversation. 
While the heuristic rewards that we defined are amenable to automatic calculation,
and do capture some aspects of what makes a good conversation, ideally 
the system would instead receive real rewards from humans.
Another problem with the current model is that we can only afford to explore a very small number of candidates and simulated turns since the number of cases to consider grow exponentially. 

\section{Conclusion}
We introduce a reinforcement learning framework for neural response
generation by simulating dialogues between two agents, integrating the strengths
of neural \sts systems and reinforcement learning for dialogue.
Like earlier neural \sts models, our framework
captures the compositional models of the meaning of a dialogue turn and 
generates semantically appropriate responses. Like reinforcement learning 
dialogue systems, our framework is able to generate utterances that 
optimize future reward, successfully capturing global properties of a good conversation.
Despite the fact that our model uses very simple, operationable heuristics for capturing these global properties, the framework generates more diverse, interactive responses that foster a more sustained conversation. 
\section*{Acknowledgement}
We would like to thank 
Chris Brockett, Bill Dolan and other members of
the NLP group at Microsoft Research for insightful
comments and suggestions. We also want to thank
Kelvin Guu, Percy Liang, Chris Manning, Sida Wang,
Ziang Xie and other members of the Stanford NLP
groups for useful discussions. Jiwei Li is supported
by the Facebook Fellowship, to which we gratefully
acknowledge. This work is partially supported by the NSF via
Awards IIS-1514268, IIS-1464128, and by the DARPA Communicating with Computers (CwC) program under ARO prime contract no. W911NF- 15-1-0462.
Any opinions, findings, and conclusions
or recommendations expressed in this material
are those of the authors and do not necessarily
reflect the views of NSF, DARPA, or Facebook.

\bibliographystyle{emnlp2016}
\bibliography{emnlp2016}

\end{document}